\pdfoutput=1

\documentclass[11pt]{article}

\usepackage[]{naacl2021}

\usepackage{times}
\usepackage{latexsym}

\usepackage[T1]{fontenc}

\usepackage[utf8]{inputenc}

\usepackage{microtype}

\usepackage{array}
\usepackage{graphicx}
\usepackage{comment}
\usepackage{tabularx}
\usepackage{blindtext, subfig}
\usepackage{dblfloatfix} 
\usepackage{multirow}
\usepackage{float}
\usepackage{boxhandler}
\usepackage{amsmath}
\DeclareMathOperator*{\argmax}{argmax}
\usepackage{amssymb}
\usepackage{environ}
\usepackage{enumitem}
\usepackage[export]{adjustbox}
\usepackage{hhline}
\usepackage{arydshln}
\usepackage{boldline, makecell}
\usepackage{esvect}
\usepackage{tcolorbox}

\usepackage{lipsum}
\usepackage{mathtools}
\usepackage{cuted}
\usepackage{booktabs}

\usepackage{draftwatermark}
\SetWatermarkText{Preprint}
\SetWatermarkScale{0.2}
\SetWatermarkAngle{0}
\SetWatermarkLightness{0.80}
\SetWatermarkVerCenter{0.04\paperheight}
\SetWatermarkHorCenter{0.09\paperwidth}

\newcommand{\fullversion}[1]{}
\newcommand{\swallow}[1]{}

\newcommand{\todo}[1]{}

\title{Hierarchies over Vector Space: Orienting Word and Graph Embeddings}

\author{Xingzhi Guo \\
  Department of Computer Science,  \\
  Stony Brook University, NY, USA  \\
  \texttt{xingzguo@cs.stonybrook.edu} \\\And
  Steven Skiena \\
  Department of Computer Science  \\
  Stony Brook University, NY, USA\\
  \texttt{skiena@cs.stonybrook.edu} \\}

\begin{document}
\maketitle
\begin{abstract}
Word and graph embeddings are widely used in deep learning applications. 
We present a data structure that captures inherent hierarchical properties from an unordered flat embedding space, particularly a sense of direction between pairs of entities.
Inspired by the notion of \textit{distributional generality} \cite{weeds2004characterising}, our algorithm constructs an arborescence (a directed rooted tree) by inserting nodes in descending order of entity power (e.g., word frequency), pointing each entity to the closest more powerful node as its parent. 

We evaluate the performance of the resulting tree structures on three tasks: hypernym relation discovery, least-common-ancestor (LCA) discovery among words, and Wikipedia page link recovery.  We achieve average 8.98\% and 2.70\% for hypernym and LCA discovery across five languages and 62.76\% accuracy on directed Wiki-page link recovery, with both substantially above baselines. 
Finally, we investigate the effect of insertion order, the power/similarity trade-off and various power sources to optimize parent selection.
\end{abstract}

\section{Introduction}
Word and graph embeddings have important applications outside their primary mission as features \cite{gillespie2020improving, lin2023comet, borca2022provable, guo2019inferring, } for machine learning models \cite{zhang2018dual, zhang2017fast2, zhang2016symmetrical} , such as word analogy \cite{gladkovaetal2016analogy} bio-informatics \cite{sultan2022low}, anomaly detection\cite{guo2022subset, zhang2023subanom, guo2022verba} and graph visualization \cite{wang2019deepdrawing}.
Embeddings generally consist of unordered points in $d$-dimensions where $d$ typically ranges from 50 to 300. But they are flat, with no explicit organization beyond geometrical proximity.   In this paper, we aim to expand the power of embeddings for both exploratory analysis and machine learning by building hierarchical structures on top of them.

Directed rooted trees (arborescences) are important structures with many natural applications.
When modeling a taxonomy or hierarchy, leaf-to-root paths typically define successively more general concepts or more powerful entities.    
Sets of nodes in small sub-trees typically share similarities and define natural clusters in the dataset.  

\begin{figure}[h]
\centering
    \subfloat[][Embedding provides similarity, but no explicit structure.]{
      \includegraphics[width=3.5cm] {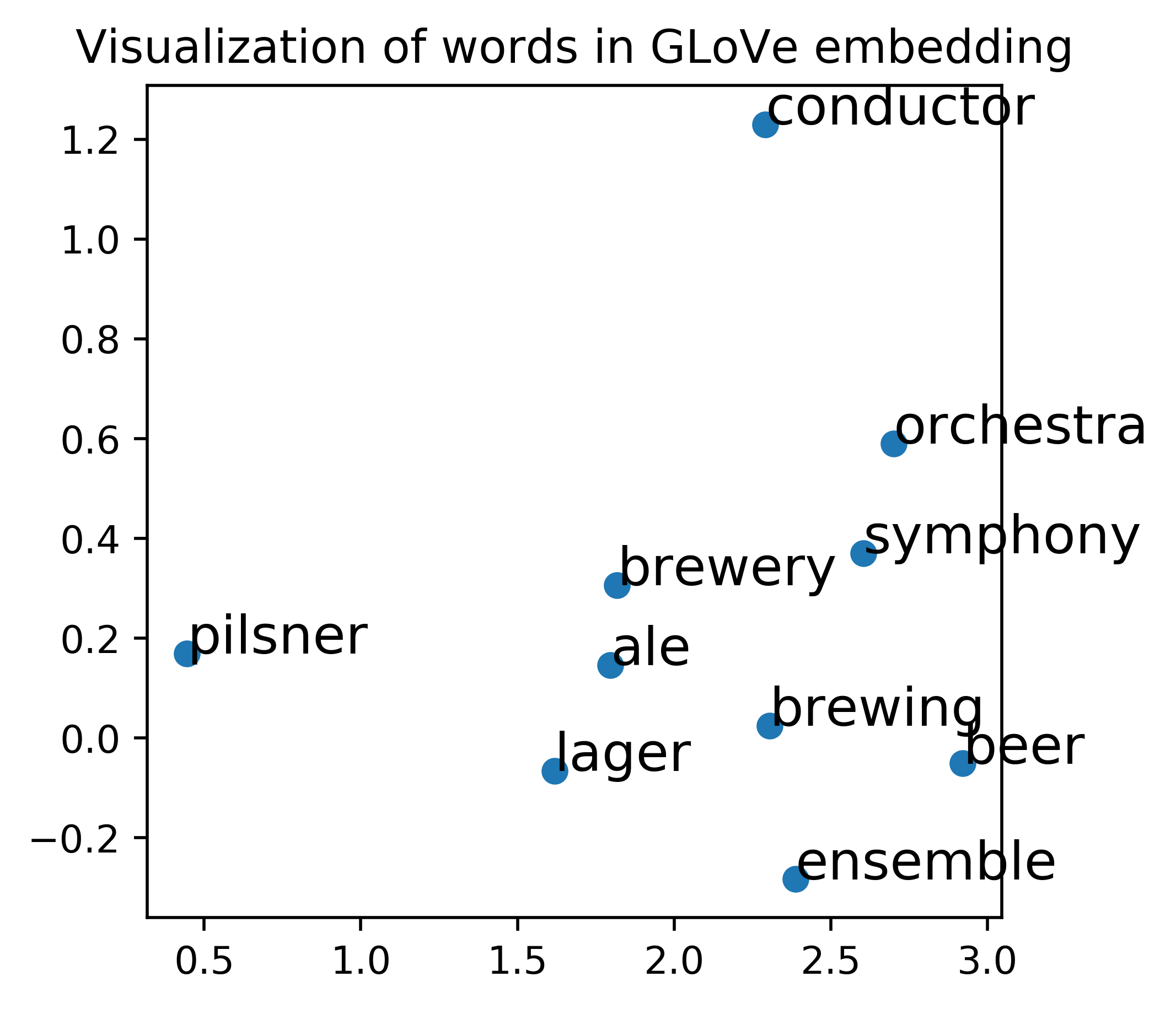}
     }\hfill
    \subfloat[][The sub-trees extracted from the embedding space. ]{
      \includegraphics[width=3.5cm, height = 3.5cm] {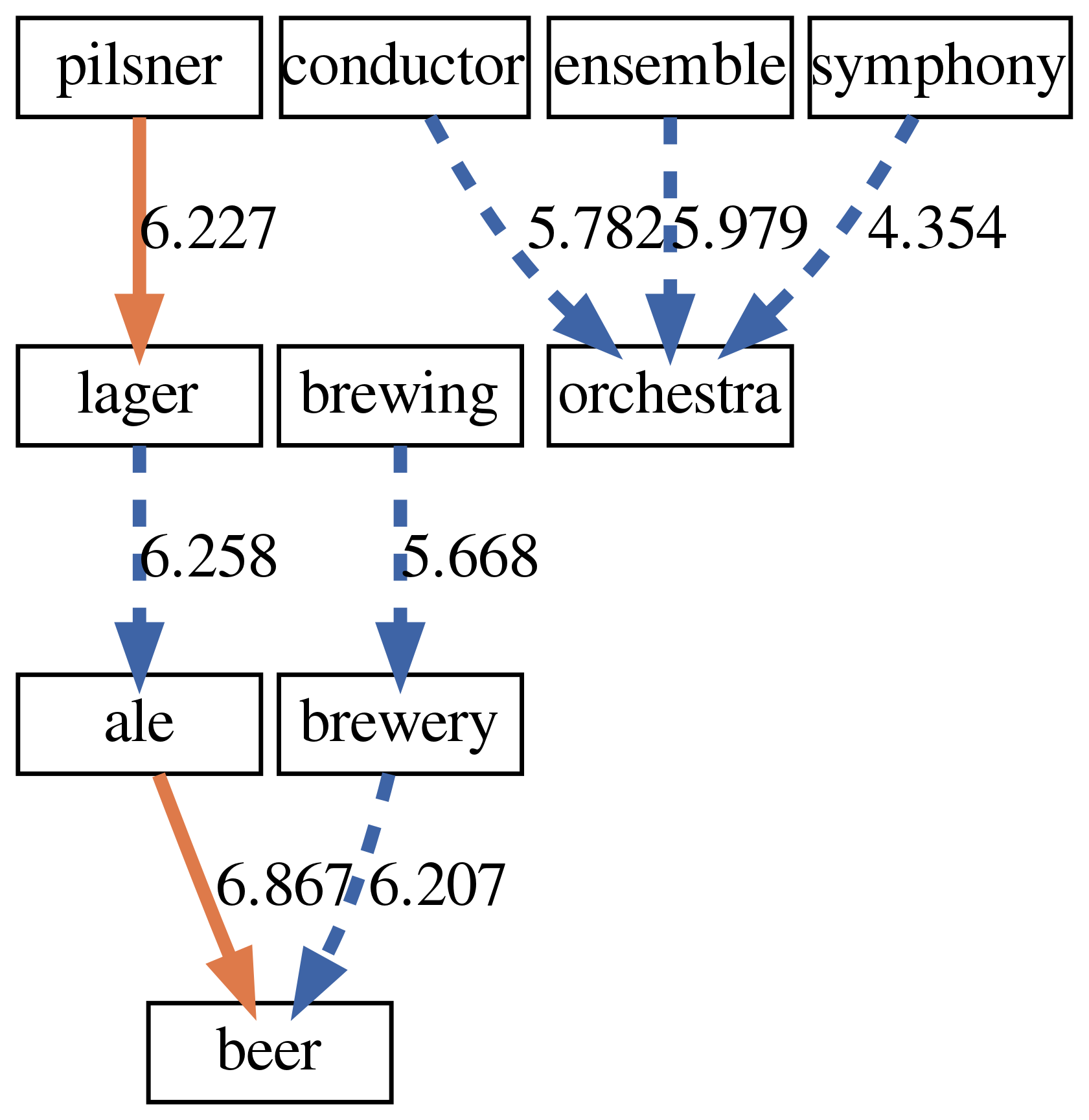}
     }
     
\caption{ Sub-figure(a): 2D PCA projection of words in GLoVe embedding. Sub-figure (b): Discovered solid edges exist in WordNet, while dot edges do not. Associated edge length reflects the $l^2$ distances between words in embedding space. By finding the directed edges among them, a meaningful hierarchy could be discovered. }
\label{illustrate_word_embedding_edges}
\end{figure}

Unfortunately, many of the current embeddings have no explicit structure. As Figure \ref{illustrate_word_embedding_edges}(a) shows, the distance metric between entities reflects similarity, but the embedding itself does not have an explicit ordered structure, despite that words and graph certainly have underlying hierarchical structures. 

In this paper, we propose a simple but powerful way to construct meaningful hierarchies, or orientations, from almost any embedding.  
We incorporate one additional feature, called \textit{entity power}, which is easily obtained at the time of building the embedding, namely a notion of the frequency, magnitude, or importance of each entity.    
For words, the word frequency of each token in the training corpus makes for a natural notion of importance, while for graphs the vertex degree of each node or its PageRank \cite{page1999pagerank, chen2023accelerating, zhou2024iterative} plays a similar role. 
We further observe the word embedding implicitly encodes word frequency information, which is consistent with observations found in \citet{schnabel2015evaluation,gong2018frage} .

Inspired from \textit{distributional generality} proposed by \citet{weeds2004characterising}, we assume that general entities have more power than the specific ones, implying a rational entity insertion order when construct arborescences over embeddings.
After starting from a single root node at the center of the embedding space, we construct our arborescence in an iterative fashion, inserting nodes in order from most to least powerful, having each new node point to its nearest neighbor already in the tree as its parent, representing similarity in edge length.

We demonstrate that the arborescences we build from word and graph embeddings have several appealing properties as Figure \ref{Tree_example} shows.    
We envision a variety of applications associated with visualization and reasoning via embeddings.
Consider the task of discovering hypernym relationships for low resource languages.  
\textit{Hypernyms} are “type-of” relations, for example, “color” is the hypernym of “red” and “blue” .  WordNet \cite{miller1995wordnet} and BabelNet \cite{navigli2012babelnet} are important sources of hypernym relations for popular languages, such resources are rare among the world’s languages.   
But the oriented embeddings we propose are readily constructed from any text corpus, and provide a first stab at discovering hypernym relationships in the absence of any additional resources.

Our primary contributions in this paper are:

\begin{itemize}
\item \textit{Constructing and Evaluating Oriented Word and Graph Embeddings}: We propose and evaluate a variety of parental selection decision rules for orienting embeddings, differing by insertion order and nearest neighbor criteria.  We demonstrate that our favored insertion ordering of most-to-least powerful proves most successful at identifying edges (in graph embeddings) and discovering hypernym relationships (in word embeddings). Figure \ref{hypernym_rank} shows the word frequency relations between hypernym/hyponyms pairs in WordNet \cite{miller1995wordnet}, suggesting that the most hypernyms are more frequent than their hyponyms, and presumably should be inserted first. 

\begin{figure}[!htbp]
\centering
    
    \subfloat[][English]{
    \includegraphics[width=3.5 cm, keepaspectratio]{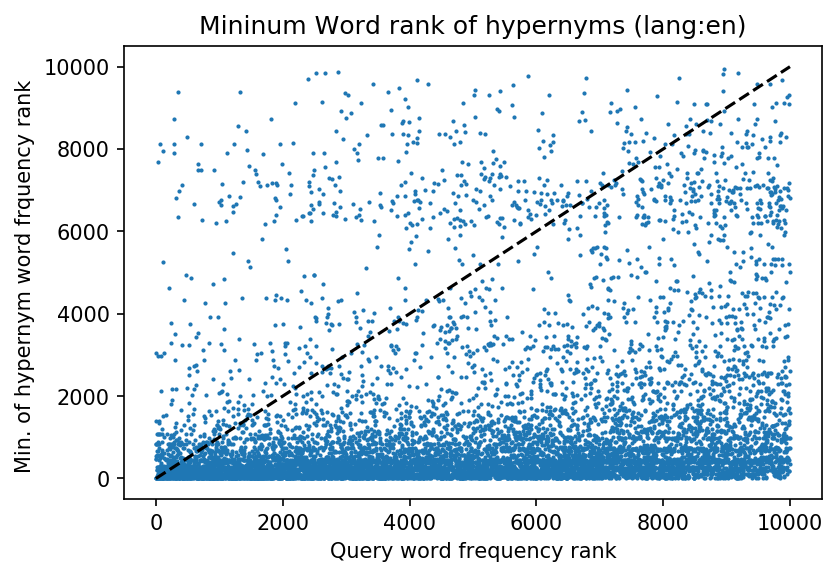}
    }\hfill
    \subfloat[][French]{
    \includegraphics[width=3.5 cm, keepaspectratio]{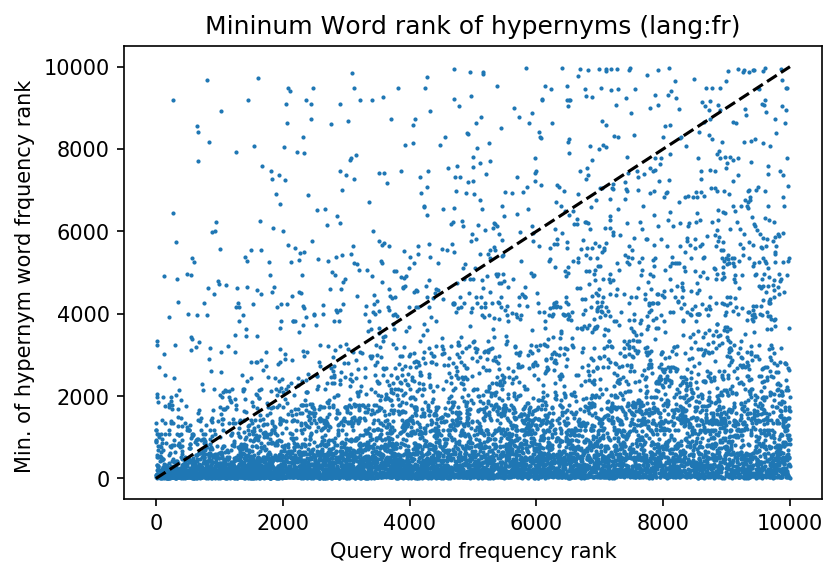}
    }\hfill
    \subfloat[][Italian]{
    \includegraphics[width=3.5 cm, keepaspectratio]{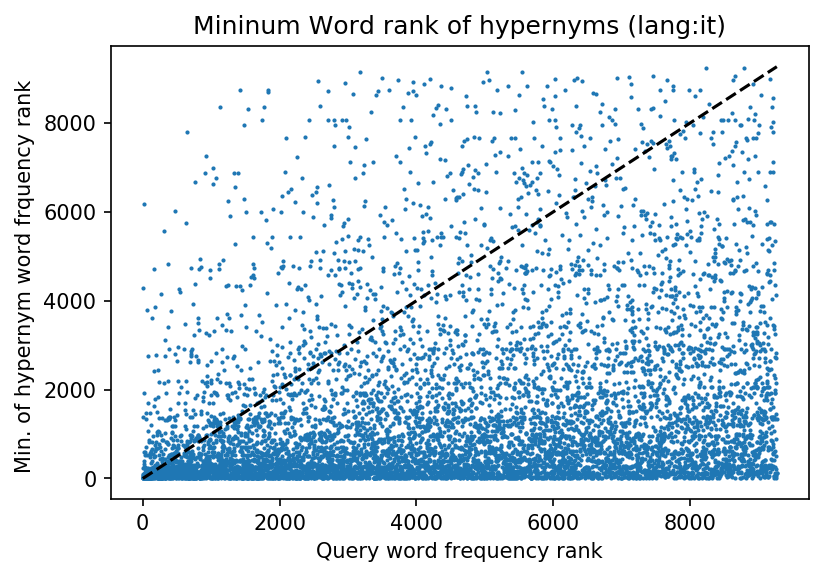}
    }\hfill
    \subfloat[][Japanese]{
    \includegraphics[width=3.5 cm, keepaspectratio]{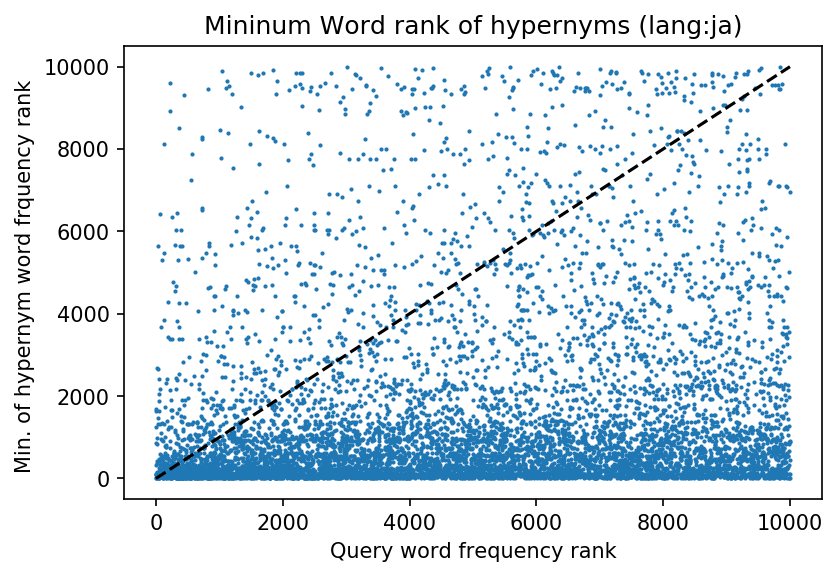}
    }\hfill

    \caption{ Plot of the word frequency rank of hypernym/hyponym word pair. Rank 0 represents the most frequent word. If multiple hypernyms exist for a word, the minimum rank of hypernym word frequency is shown. The horizontal and vertical axis represent the word frequency rank of hyponym (e.g., \textit{banana}) and hypernym (e.g., \textit{fruit}) respectively.  Most of the blue dots are placed below the black diagonal line, indicating that the most hypernyms are more frequent than their hyponyms.}
  \label{hypernym_rank}
 \end{figure}

\begin{figure*}[!htbp]
\centering

    
  \hspace*{-0.9cm}\subfloat[][Wiki: \textit{Elvis Presley}]{
    \includegraphics[width=3.5cm, keepaspectratio]{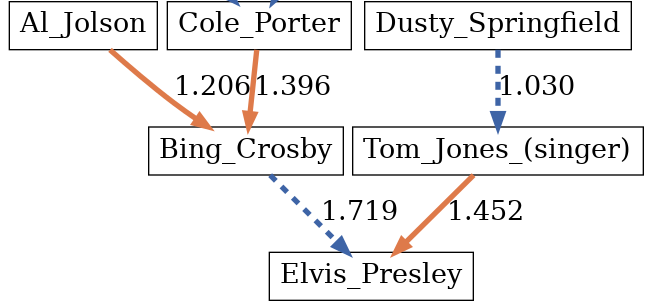}
 }
 \subfloat[][Wiki: \textit{Richard Nixon}]{
    \includegraphics[width=3.7cm, keepaspectratio]{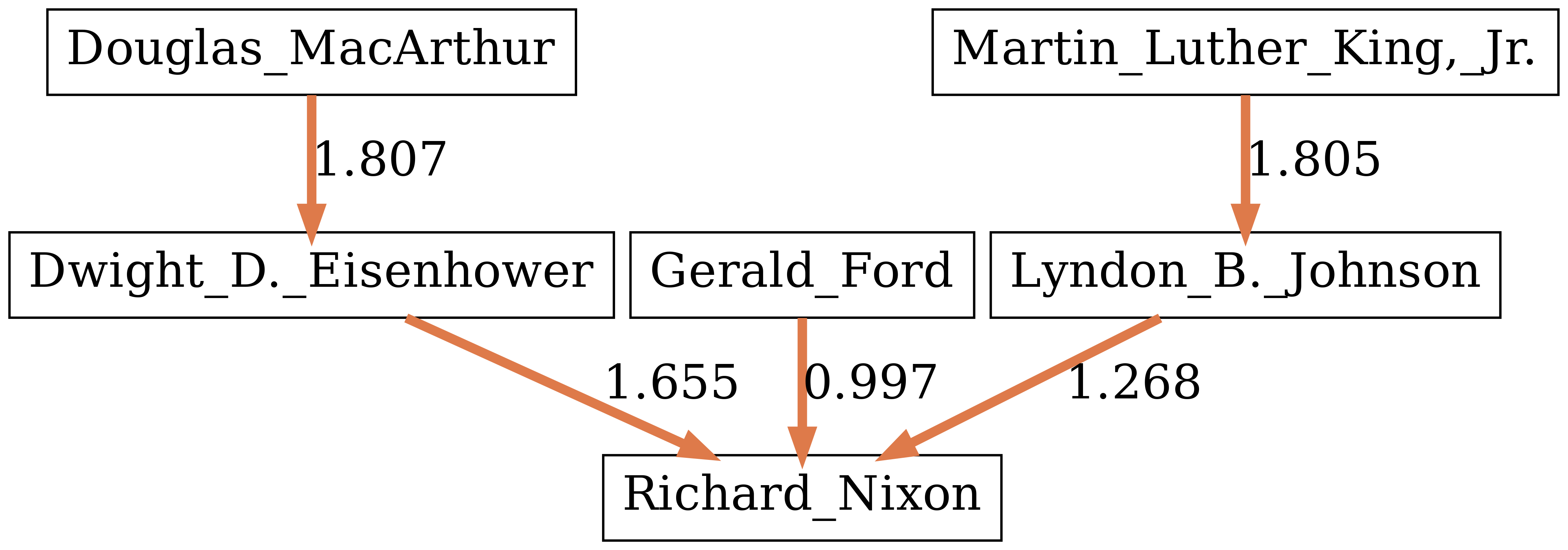}
 }
 \subfloat[][WordNet: \textit{fruit}]{
    \includegraphics[width=3cm, keepaspectratio]{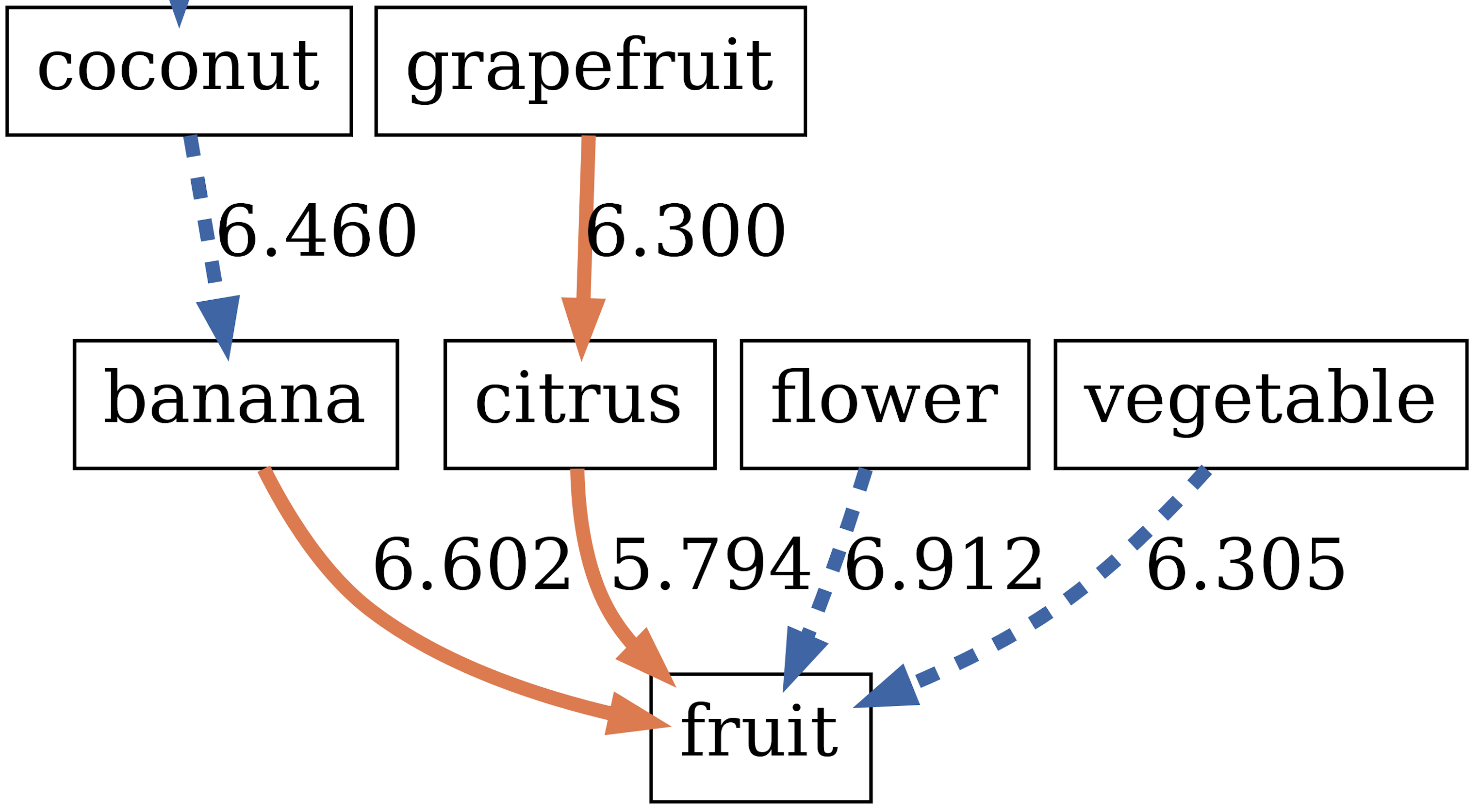}
 }
 \subfloat[][WordNet: \textit{fabric}]{
    \includegraphics[width=3cm, keepaspectratio]{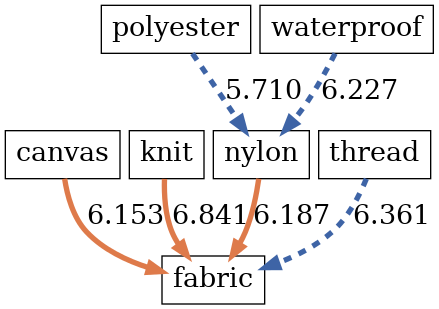}
 }
 \subfloat[][WordNet: \textit{mathematics}]{
    \includegraphics[width=4cm, keepaspectratio]{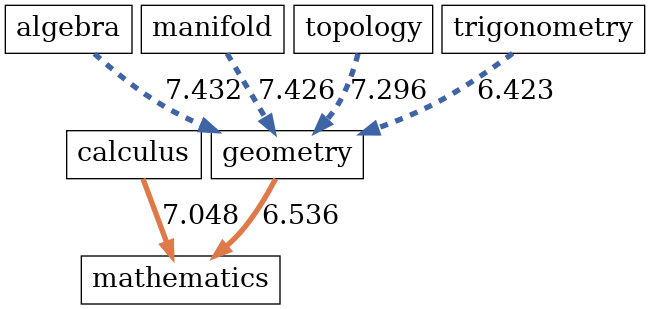}
 }
\caption{Examples of sub-trees capturing similarity/hypernym relationships.  Solid edges reflect the relationships overlapped with ground truth, dot edges indicate the discovered links not existing in the ground truth, with associated $l^2$ distances reflecting the degree of similarity. }
\label{Tree_example}
\end{figure*} 

\item \textit{Identifying Semantic Breaks}: Leaf-to-root paths repeatedly transition in and out of natural neighborhoods to more general concepts.
We demonstrate that our accuracy of identifying relationships increases when we exclude natural semantic breaks formed by long edges in the arborescence.    
This provides natural ways of efficiently constructing fine-grain clusters of entities. 


\item \textit{Entity Power vs. Similarity in Identifying Relationships}: Nearest neighbor models are widely used in NLP and data science to ascribe possible relationships between pairs of items.  But power law distributions imply relatively few entities will account for an outsized fraction of total references or interest.   
Inspired from \textit{Distributional Generality Hypothesis} \cite{weeds2004characterising}. We incorporate notions of power into similarity detection seems valuable in breaking near-ties and prioritizing attention. We perform a series of experiments of the performance of distance measures incorporating various power components, establishing the best trade-off between them.

\item \textit{Hidden Power Source from Word Vector: }: The frequency rank might be inaccessible from pure word embedding matrix. Fortunately, from our experiment and recent research \cite{gong2018frage, arora2016latent}, we confirm that the first few principal components (PCs) encode the word frequency information. Leveraging this finding, our experiments demonstrate that the power order induced from PCs is significant better than random and helpful in tree construction.

\item \textit{Algorithmic Implications of Hierarchical Structures}:
For example, the lowest common ancestor (LCA) \cite{schieber1988finding, bender2001finding} of nodes $x$ and $y$, represents the unique tree node $l$ most distant from the root that is an ancestor of both $x$ and $y$, typically defining the commonalities or superordinate relations between them. 
After linear-time pre-processing, the lowest common ancestor of $(x,y)$ can be computed in constant time for any node pair, providing a fast way to compare embedded representations. 
\end{itemize}

\section{Related Work}

Word \cite{mikolov2013distributed,pennington2014glove} and graph embeddings \cite{perozzi2014deepwalk} are widely used to capture the relevance of symbolic entities in a lower-dimensional space, comparing to the conventional metrics such as Personalized PageRank \cite{zhou2024iterative, chen2023accelerating}.

Rooted on Distributional Hypothesis \cite{harris1954distributional} - the entities in the similar contexts tend to have similar meaning. 
The co-occurrence based embedding methods capture the correlation of semantic relations. \citet{schnabel2015evaluation} applied intrinsic evaluation on word embeddings and showed that the nearest neighbors of a queried word highly overlap with related words labeled by human (e.g., {\em money} is close to {\em cash}).


Natural languages inherently have hierarchical structures \cite{gaume2006hierarchy,polguere2009lexical, wang2022knowledge, wang2023knowledge, wang2023knowledge2}. Distributional Generality was proposed by \citet{weeds2004characterising}, suggesting contexts of a specific word (hyponym) are included in its semantically-related but more general word (hypernym), and hypernyms tend to be more frequent than the hyponyms. \citet{hearst1992automatic} leveraged the text surface pattern to discover hypernyms and extended text patterns by bootstrapping. \citet{snow2005learning} further built a classifier for hypernym word pair detection in a vector-space model. \citet{herbelot2013measuring} proposed \textit{Semantic Content} and used KL divergence as the hypernym score, but it does not outperform a simpler frequency measure. 




As word embedding and deep learning become popular, many researchers aim at building embedding model dealing with hierarchical information.
\citet{alsuhaibani2018joint} proposed a hierarchical word embedding model considering not only the contextual information, but also hypernym specific information on a taxonomy for fine-tuning. \citet{fu2014learning},\citet{vylomova2015take}, \citet{weeds2014learning} and \citet{rimell2014distributional} detect hypernym relations in a supervised word pair classification framework based on the geometric properties of embeddings such as clustering, embedding negation and other vector operations. Recently, \citet{nickel2017poincare,tifrea2018poincar} proposed to model tree structure in hyperbolic space, preserving tree-distance in non-Euclidean Poincaré space, and leverage the resulting embedding geometry to reconstruct word hierarchy and predict missing links.

Unfortunately, these aforementioned approaches need labeled structured data (e.g., hypernyms from WordNet \cite{miller1995wordnet}), word pairs with known relations or hand-crafted patterns, and they may not be easily expandable to resource-poor languages. Different from the previous works, our goal is \textbf{\em not} to design a supervised model for word pairs classification, but to discover a meaningful hierarchy over embeddings of symbolic data in an unsupervised manner, then evaluate the hierarchy's rationality with external ground truth.



\section{Methods}





We propose an unsupervised method to build an arborescence over symbolic data, enhancing embeddings with a measure of the power or importance of each data point, like word frequency or vertex degree. 
After initializing the tree with a maximally powerful root node artificially, simulating the ultimate parent of all entities, at the center of the embedding space, we insert entities into the tree in descending order of entity power.

Each new node is connected to the closest current tree node as the parent by some measure of similarity.   
Here we evaluate a family of similarity measures based on geometric distance (Euclidean or cosine distance) in embedding space and node power of the parent candidates.
With the selected parent $P_i$ for the $i^{th}$ inserted node, we defined the similarity measures below:
\begin{align*}
P_i = \argmax_{Node_j\in  T_{i-1}} p*\frac{dist^{-2}(\vec{v}_i, \vec{v}_j)}{Max(dist^{-2}(\vec{v}_i, \vec{v}_{k \in \{ 0,\dots,j\} }))}  \nonumber \\ 
+ (1-p)*\frac{log(d_j) }{Max(\log(d_{k \in \{ 0,\dots,j\} }))}     
\label{eq:1}
\end{align*}
%
where $T_{i-1}$ is the total tree nodes prior to the new node being inserted, $p \in [0, 1] $ is the weight on distance,  $d_j$ is the entity power of the candidate parent $Node_j$, $\vec{v_i}$ is the embedding vector of $Node_i$, and $dist(\cdot)$ is either $l^2$ or cosine distance function.

The range of edge scores is scaled to be $[0,1]$ by separately normalizing the contributions from node distance and node power, and then linearly combining them according to the weight $p$. In this manner, the edge score is proportional to the square root of node distance and logarithm of node power. 

Meanwhile, to avoid the power dominance of the artificial root, we dynamically change root's power to be the average power of the inserted entities. Therefore, the first inserted nodes are more prone to connect the root, preserving the natural high-level categorization or semantic base. As insertion continues, the root becomes less influential.


Inserting each of the $n$ nodes requires searching for the nearest neighbor among all current tree nodes, so the complete tree can be constructed in $O(n^2)$ time. The bottleneck here is in finding the nearest neighbor. 
Instead of using brute force, a geometric search data structure such as a ball-tree \cite{omohundro1989five} can be pre-computed over all the points.   
Although we are only interested in the nearest more-powerful point instead of the closest point over all, we can heuristically reduce the search space using ball-trees, so the expected time complexity can be reduced to $O(n^{3/2})$.

\begin{figure}[!htbp]
\centering
      \includegraphics[width=6 cm, height= 4 cm]{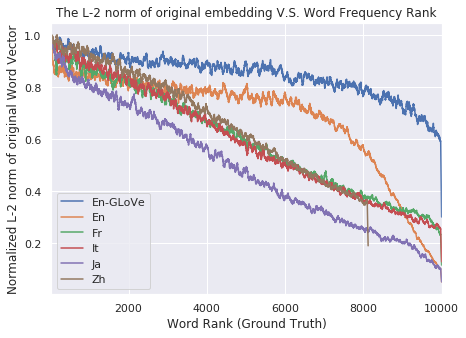}
  \caption{We select the vocabularies overlapped with the WordNet, roughly 10K for each language. Word frequency decreases from left to right along horizontal-axis. Concurrently, the $l^2$-norm of the word vector tends to decrease, indicating a strong correlation between word frequency and vector length. The value is smoothed using a window of size 50. }
  \label{L2norm_word_freq}
 \end{figure}

Besides the explicit word frequency counted from the corpus, \citet{arora2016latent} theoretically explains the relation between word frequency and $l^2$-norm of the word vector, showing that high frequency words tend to have large $l^2$-norm, as we observed in Figure \ref{L2norm_word_freq}

Based on this justification, \cite{mu2017all} proposed a simple but effective post-processing method for word embedding, arguing that the undesirable word frequency is encoded in the first few principal components (PCs) of word embedding. 
Their post-processed embedding, by projecting PCs away from word vectors, presumably removes the frequency bias, and gain better performance consistently on various word similarity tasks. 

Learning from their investigations, we leverage the $l^2$-norm of the projected PCs as a new source of word power, called \textit{PCA-induced Power}. Formally:
\begin{equation}
    \begin{aligned}
    u_0, u_1, ..., u_d &= PCA(V - \mu) \\[3pt]
    d_w &= ||\sum_{i=0}^{k}{ (v_w u_i^{\intercal}})u_i||_2 \\[3pt]
    \widetilde{v}_w&= v_w - \mu - \sum_{i=0}^{k}{ (v_w u_i^{\intercal}})u_i
    \end{aligned}
    \label{pca_equation}
\end{equation}

where $V$ is the word embedding matrix, $\mu$ is the center of the embedding vectors, $u_i$ is the $i^{th}$ principal component of the embedding centered at $\vv{0}$, $d_w$ is the PCA-induced power for the word $w$, $v_w$ is the row-vector embedding of the word $w$, $\widetilde{v}_w$ is the de-biased word vector, $k$ is the parameter controlling how many PCs should be projected. We fix $k$ to be 3 for all experiments.

\newcommand{\treeWidth}{3.9}
\newcommand{\treeHeight}{3}

\section{Experiments}
We construct our proposed trees for two types of embedding datasets:
the Wikipedia-people graph embedding and word embeddings of multiple languages.  We investigate effects of three node insertion orders (descending, random and ascending), the power/distance trade-off parameter $p$, and the word power induced from word vector.  
In the following section, we will describe the datasets, report our results on Wiki-people and WordNet (English) in detail, then extend our experiments to WordNet of other languages (French, Italian, Japaneses and Chinese), finally discuss the effect of word power information encoded in the word vector.

\subsection{Dataset Description}

\begin{table*}[htb!]
    \centering
    \begin{tabular}{  m{1.75cm}  m{2.7cm} m{1.8cm} m{1.8cm} m{1.8cm} m{1.8cm} m{1.8cm} m{1.8cm} } 
    \Xhline{1pt}
    Dataset &Wikipedia People  &WordNet(En) &WordNet(Fr) &WordNet(It) &WordNet(Ja) &WordNet(Zh) \\ 
    \hhline{=======}
    \#Nodes &100,000 &10,000 &10,000 &10,000 &10,000 &8,188    \\ 
    \hdashline
    \#Edges  & 441,815 & 223,817 & 188,754 & 110,859 & 150,331 & 62,854   \\ 
    \hdashline
    Embeddings  & Deepwalk & GLoVe, Word2Vec & \multicolumn{4}{c}{ Word2Vec }   \\ 
    \hdashline
    Power Measure  & Vertex degree & \multicolumn{5}{c}{Zipf-induced Word Frequency \& PCA-induced Power} \\
    \hdashline
    Evaluation Criteria  & Link recovery & \multicolumn{5}{c}{Hypernym discovery \& Least-Common Ancestor discovery}  \\ 
    \Xhline{1pt}
    \end{tabular}
    \caption{We use CBOW version of Word2Vec in all experiments. Since WordNet (Fr, Zh, It, Ja) only share a small portion of vocabulary (\textasciitilde 10K) with the pre-trained embedding, we limit the size of all WordNet datasets to the similar scale. Multilingual WordNet data is organized by \citet{bond2013linking,bond2012survey}. The versions of each language are created by \citet{_Fellbaum:1998,Wang:Bond:2013,Isahara:Bond:Uchimoto:Utiyama:Kanzaki:2008, Toral:Bracale:Monachini:Soria:2010, Sagot:Fiser:2008} }
    \label{tab:data_stat}
\end{table*}

The Wiki-people dataset is a directed graph of links between biographical Wikipedia pages \cite{snapnets,skiena2014s, guo2021subset}. 
We sampled a sub-graph of 100K nodes from the largest weakly-connected component of the Wiki-people graph. Then we run Deepwalk \cite{perozzi2014deepwalk} on the undirected version of the sub-graph to obtain the node embeddings. We expect our algorithm can recover the directed relations from the embeddings trained from an undirected hierarchy.

WordNet \cite{miller1995wordnet, bond2013linking} contains a directed graph of words in major languages, where certain edges represent hypernym relationships, specifically, the hypernyms within a five-closure of a word.\footnote{We also did experiments with the hypernym closures from 1 to 4, with similar results.}
For example, \textit{President} is \textit{Man}, \textit{Man} is \textit{Vertebrate}, so \textit{President} is a \textit{Vertebrate}. We leverage pre-trained English GLoVe embedding \cite{pennington2014glove} with 100-dimension, and Word2Vec \cite{mikolov2013efficient} embeddings of five languages with 300-dimension. All the embeddings are pre-trained on Wikipedia data, without any explicit hypernym information. 

We restrict attention here to the intersection between embedding vocabulary and the words involved in WordNet hypernym relations.
The embeddings are ordered by frequency rank, so we estimate the word power as underlying word frequencies using Zipf's law \cite{li1992random}, called \textit{Zipf-induced frequency}. 
Furthermore, we investigate the effect of word power directly induced from Principal Components of word embeddings, mentioned earlier as \textit{PCA-induced power}. 

The gross statistics of both data sets are presented in Table \ref{tab:data_stat}. 
Figure \ref{Tree_example} shows several examples of generated sub-trees. Our algorithm recovers edges (solid lines) in the datasets and discovers new edges (dotted lines), while the edge length represents the inter-node distance.

\subsection{Detailed Analysis on Wiki-people and WordNet (En)}
In this section, we evaluate our constructed trees by edge accuracy:  what percentage of the reported tree edges are ground-truth relationships in the associated dataset. 
We investigate the effect of insertion orders and varying $p$, as well as the edge accuracy from the perspective of edge length\fullversion{, tree levels} and node power. In all evaluations, we exclude the edges linking to the artificial root.

\subsubsection{Edge Accuracy \textit{vs.}  Power/Distance}
\begin{figure}[h]
\centering
    \subfloat[][Wiki-people]{
      \includegraphics[width=\treeWidth cm, height=\treeHeight cm]{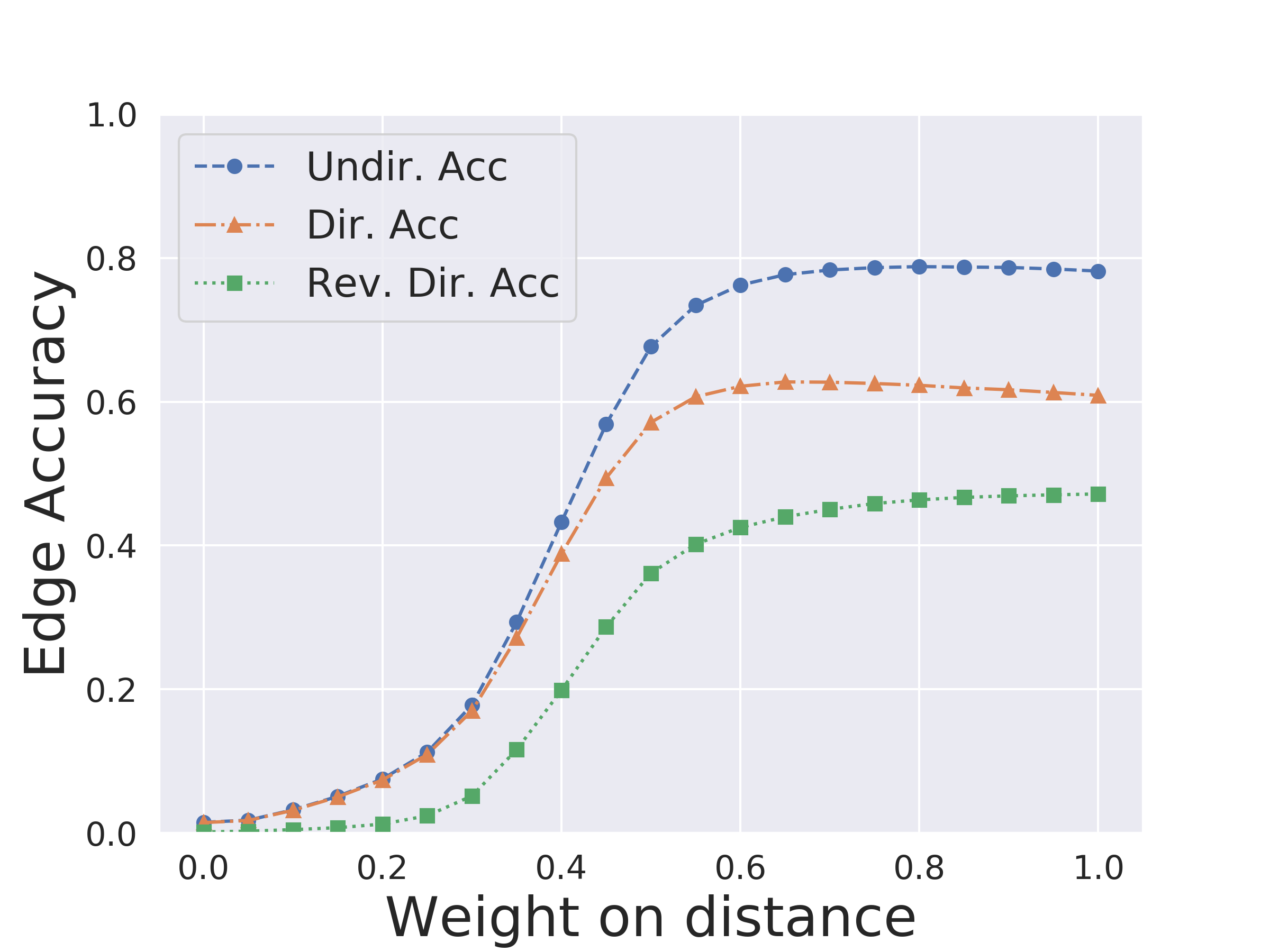}
     }
    \subfloat[][WordNet]{
      \includegraphics[width=\treeWidth cm, height=\treeHeight cm]{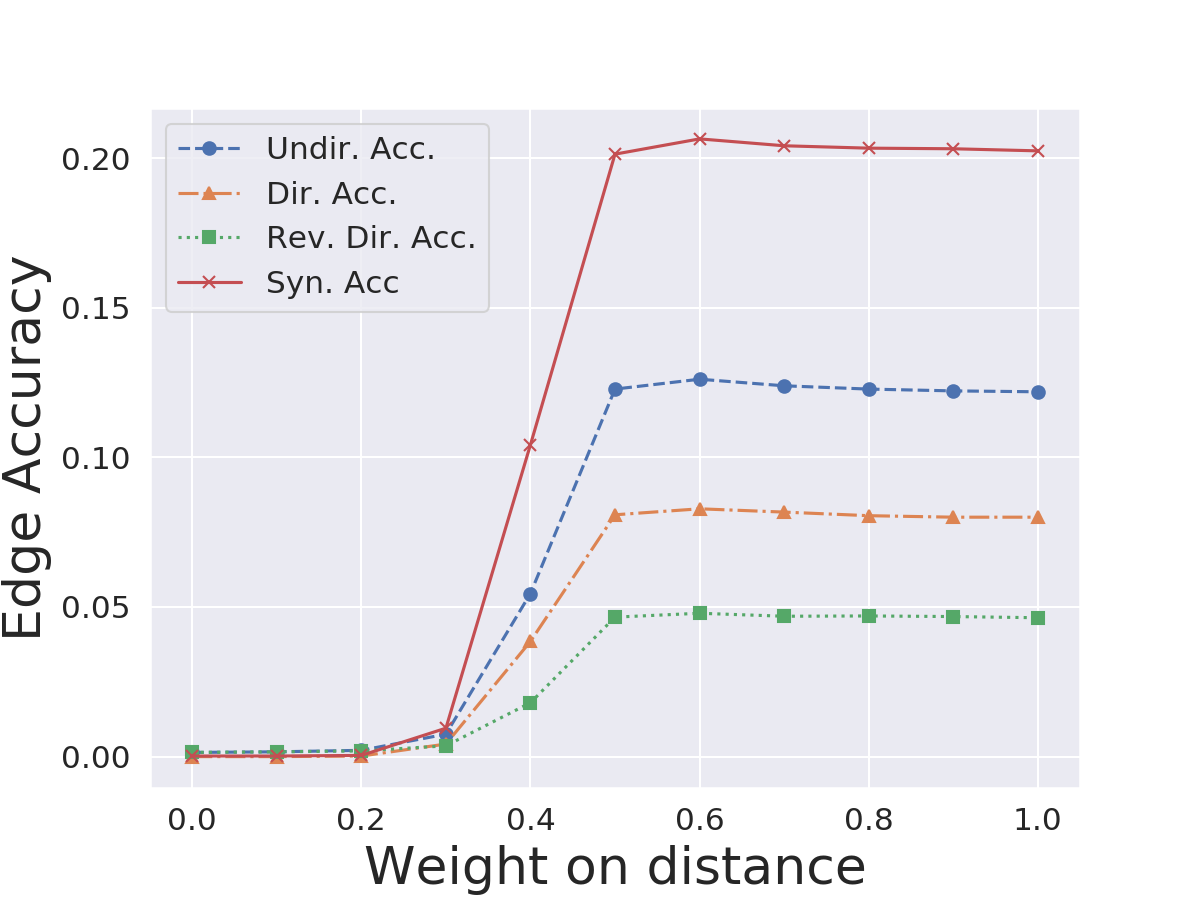}
     }
 \caption{Accuracy as $0 \leq p \leq 1$.
 Increased weight on distance over power results in trees with more qualified edges. \textit{``Syn. Acc.``} reflects the edges capturing synonym word pairs.}
 \label{weight_Trade-off}
\end{figure}

Figure \ref{weight_Trade-off} presents edge accuracy as a function of $p$, the trade-off parameter between power and distance. We measure this accuracy in three ways: {\em undirected} edge accuracy accepts edges regardless of direction, {\em directed} edge accuracy requires getting the edge orientation to match that of the ground-truth relation, and {\em reverse directed} edge accuracy counts the edges having reversed direction to the ground truth.

As expected, the edge accuracy generally increases as $p$ increases, but peaks at $p=0.6$.  This shows the importance of distance over power. 
Further, under our preferred descent insertion order the accuracy of the directed edges (\textit{`` Dir. Acc.``}) is consistently higher than the reversed directed edges (\textit{``Rev. Dir. Acc.``}), which supports our hypothesis that nodes of less power should point to more powerful ones.


Table \ref{tb:weight_Trade-off} presents model performance using the best $p$ for three different insertion orders, as well as a random selection baseline model that links to completely arbitrary parent nodes.
The model using the descending insertion order significantly outperform the others in both undirected and directed edge accuracy. 
The WordNet-Random wins for reverse directed accuracy is a Pyrrhic victory because it puts edges in backwards, but still smaller than the accuracy (0.0507 to 0.0828) of our preferred insertion heuristic.  The extremely low performance of the random selection baseline demonstrates the complexity of the edge linking task.

\newcommand{\accWidth}{1.1}
\begin{table}[!ht]
      \centering
        \begin{tabular}{ m{1.7 cm} p{\accWidth cm} p{\accWidth cm} p{1.4 cm}}
        \Xhline{1pt}
        \centering Insertion Order & Undir. Acc. & Dir. Acc. & Rev. Dir. Acc. \\
        \hhline{====}
        \centering  &\multicolumn{3}{c}{Wiki-people} \\ 
        \hhline{====}
        \centering Descent & \textbf{0.7880} & \textbf{0.6276} & \textbf{0.4717} \\
        \hdashline
        \centering Random & 0.6318 & 0.4737 & 0.4034 \\
        \hdashline
        \centering Ascent & 0.4784 & 0.3163 & 0.3565\\
        \hdashline
        \centering Random Selection & 0.0020 & 0.0009 & 0.0011 \\
        \hhline{====}
        \multirow{2}{=}{} & \multicolumn{3}{c}{WordNet(En)} \\
        \hhline{====}
      
        \centering Descent & \textbf{0.1261} & \textbf{0.0828} & 0.0479 \\
        \hdashline
        \centering Random & 0.112 & 0.0680 & \textbf{0.0507} \\
        \hdashline
        \centering Ascent & 0.0956 & 0.0482 & 0.0497\\
        \hdashline
        \centering Random Selection & 0.0021 & 0.0014 & 0.0007 \\
        \Xhline{1pt}
        
        
    \end{tabular}
    \caption{The best edge accuracy for different insertion orders.  Insertion by descending power outperforms others in both \textit{Un/directed} edge accuracy. }
    \label{tb:weight_Trade-off}
    
\end{table}

\subsubsection{ Edge Accuracy \textit{vs.} Edge Length}

\begin{figure}[!htbp]
\centering
    \subfloat[][Wiki-people]{
      \includegraphics[width=\treeWidth cm, height=\treeHeight cm]{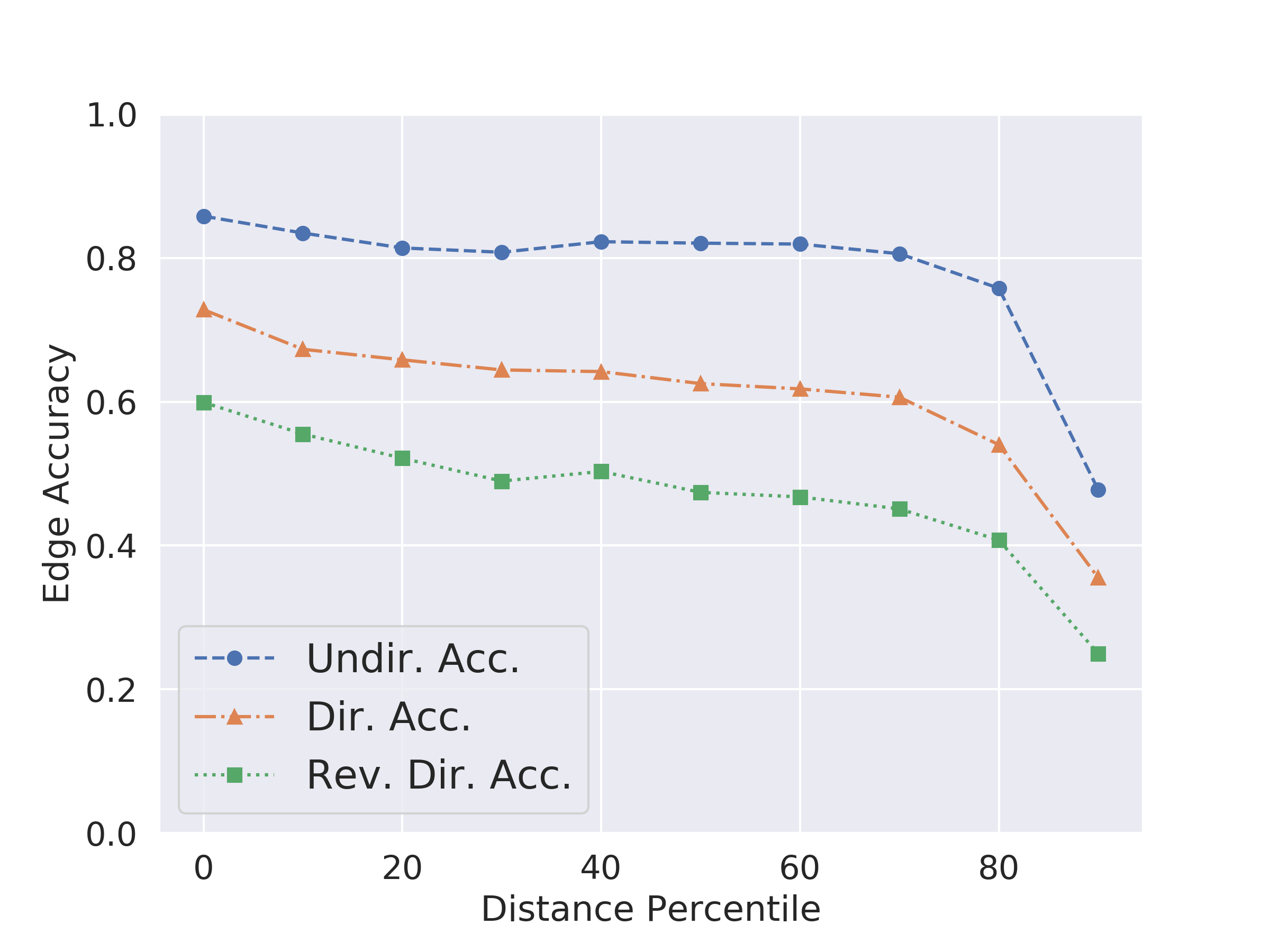}
     }
    \subfloat[][WordNet]{
     \includegraphics[width=\treeWidth cm, height=\treeHeight cm]{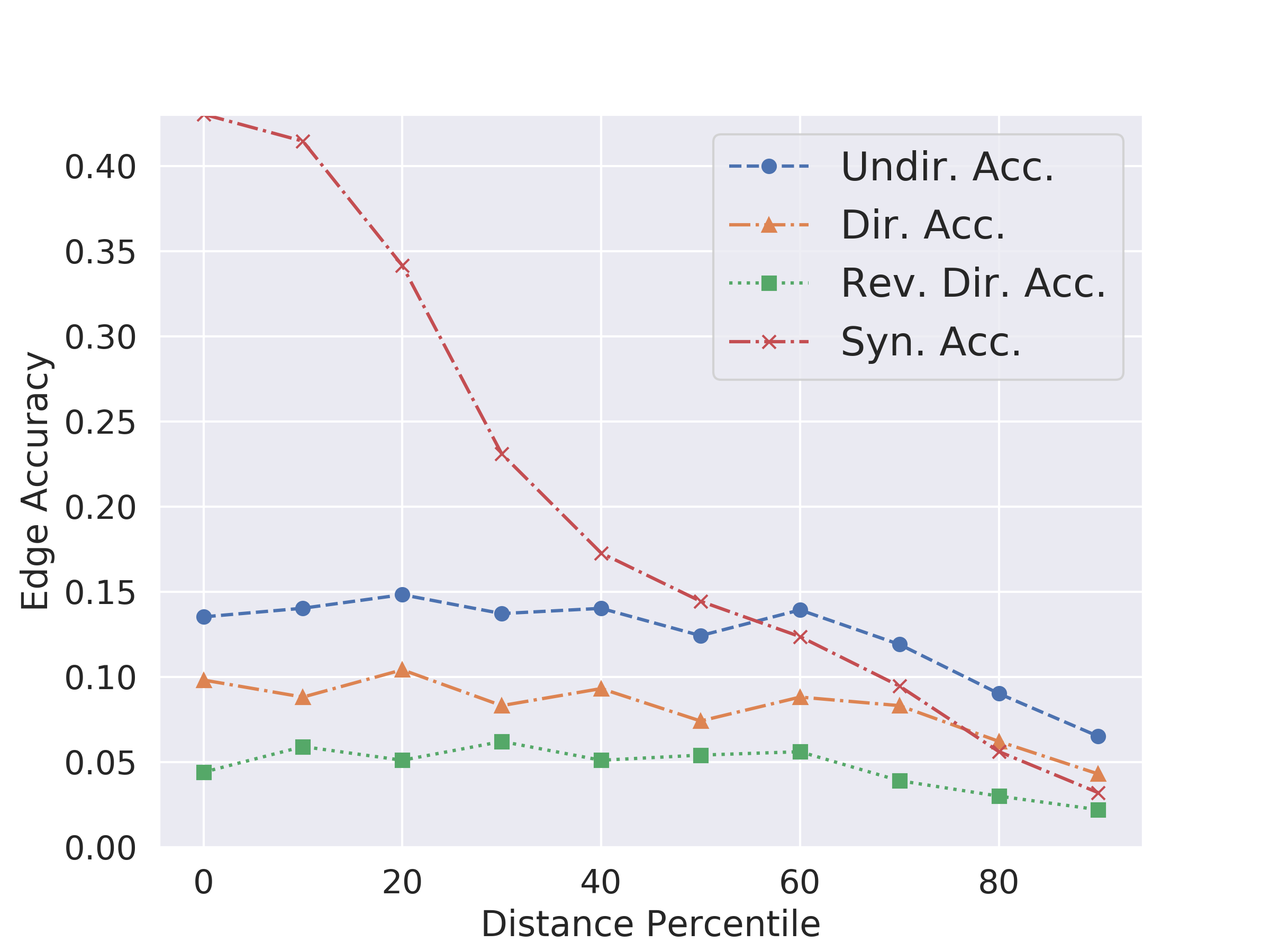}
     }
  \caption{Accuracy by edge distance. Edge with smaller distance (low percentile) is more informative, while those closest words are likely to be synonyms instead of hypernym.}
  \label{edge_distance_Trade-off}
\end{figure}

Figure \ref{edge_distance_Trade-off} reports the accuracy of tree edges as a function of the embedding distance between points (Edge length).
It shows that closer pairs generally have higher edge accuracy, matching our expectation that close embedding pairs are more suggestive of genuine relationships than distant pairs. The initial low accuracy in WordNet reflects that many of the closest words are synonyms, not hypernyms.
Generally short edges have higher accuracy, preserving more semantic and hierarchy information. Long edges create semantic breaks, forming weak links between entities. More fine-grain sub-trees can be extracted by excluding those long edges. 

\subsubsection{ Edge Accuracy \textit{vs.} Node Power }
\begin{figure}[!htbp]
\centering
    \subfloat[][Wiki-people]{
      \includegraphics[width=\treeWidth cm, height=\treeHeight cm]{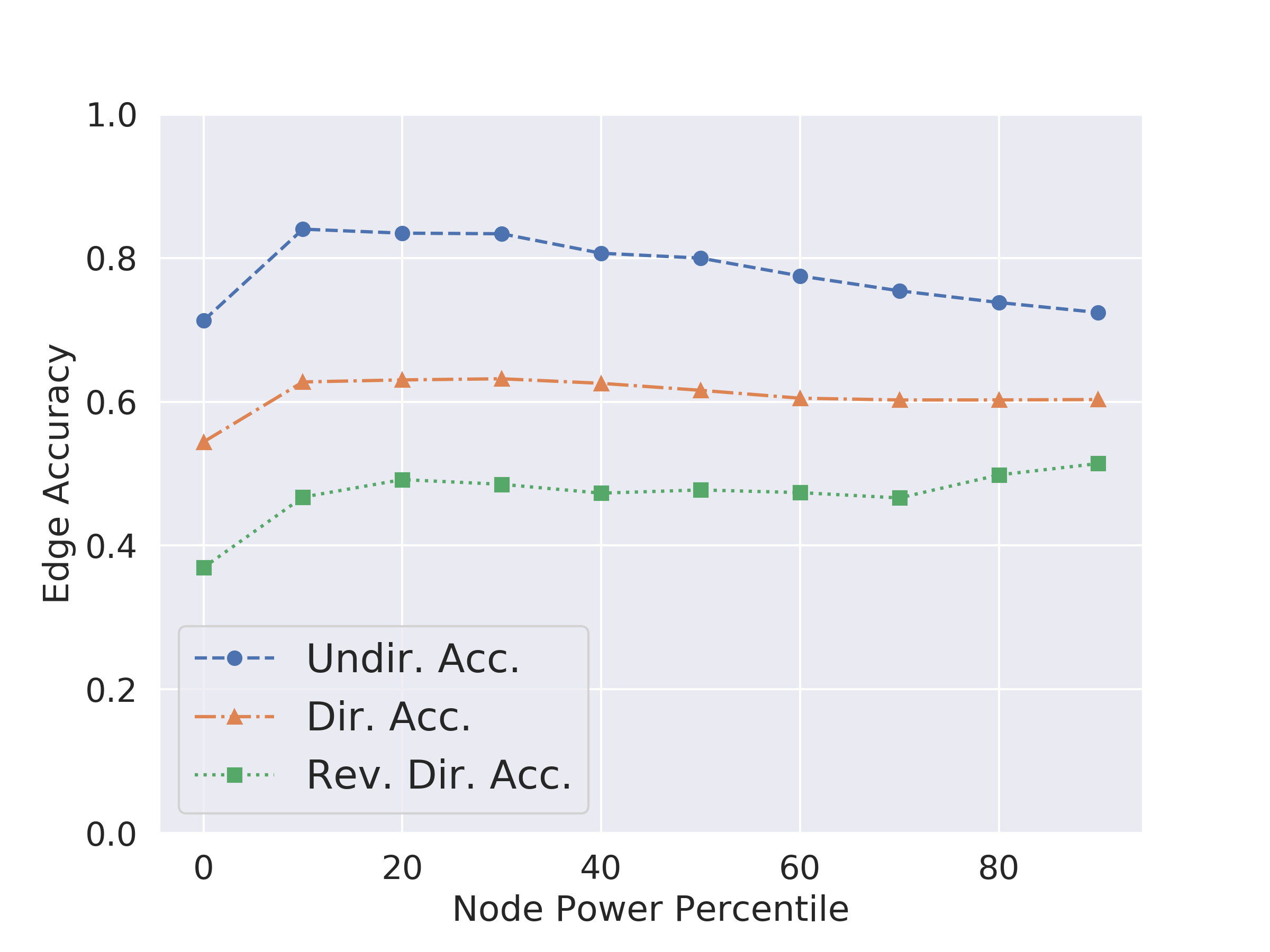}
     }
    \subfloat[][WordNet]{
     \includegraphics[width=\treeWidth cm, height=\treeHeight cm]{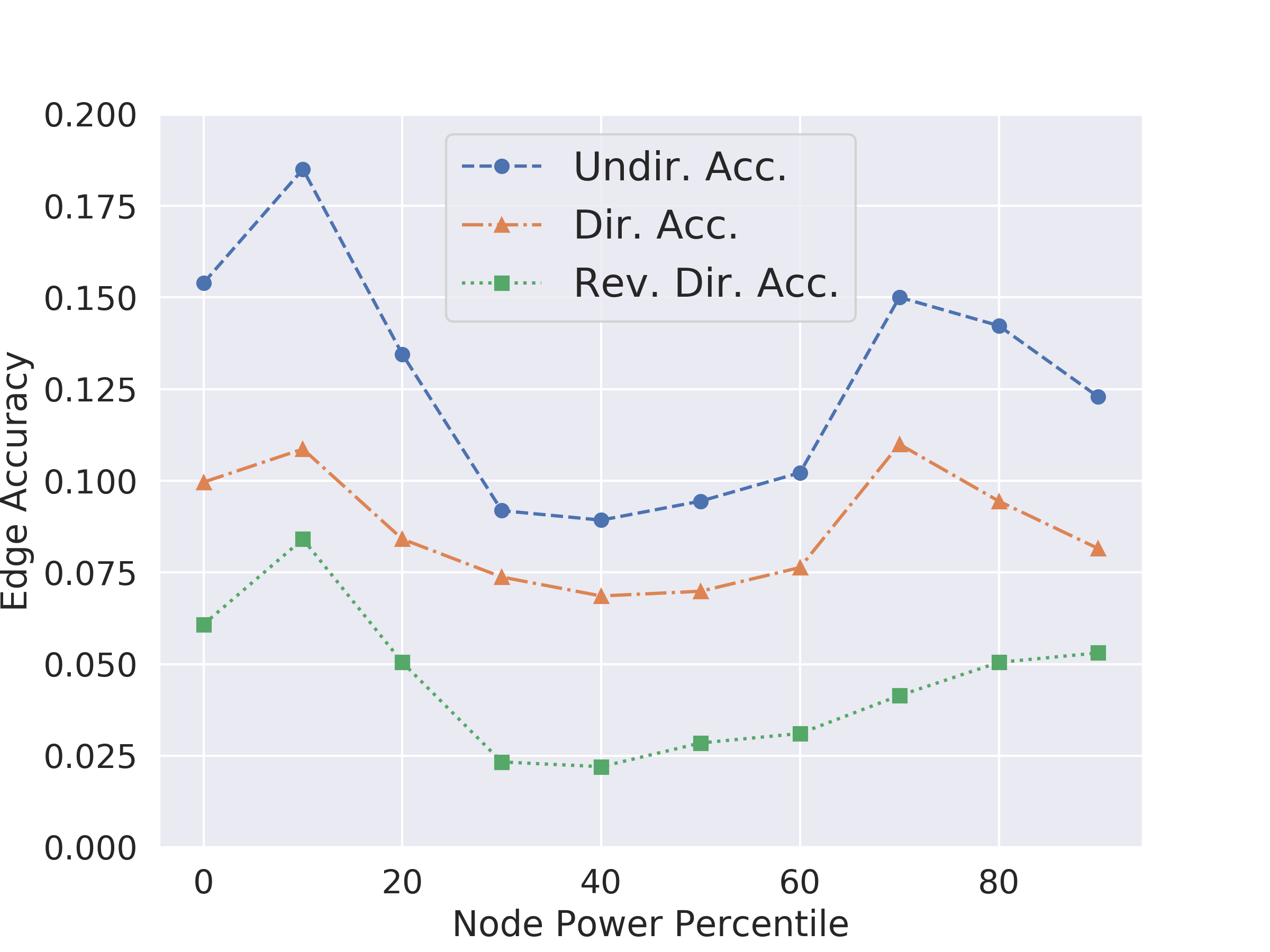}
     }
  \caption{Accuracy by node power. Under the preferred descending insertion order (from high percentile to low percentile), the directed edge accuracy increases and keeps steady.}
  \label{Bigness_Trade_off}
 \end{figure}

Figure \ref{Bigness_Trade_off} reports the edge accuracy as a function of node power. 
With the preferred insertion order, high power (of high percentile) entities are first inserted.
The Wiki-people result shows the overall accuracy keeps steady, while the low accuracy of the least powerful nodes represents the lack of community linkages in the least significant persons.
From right to left, the WordNet result shows that directed edge accuracy initially increases as we keep inserting words, demonstrating the ability to capture hypernym relations after constructing a semantic base above the artificial root.

\subsection{Multilingual WordNets and PCA-induced Word Power  }

Table \ref{multiwordnet_res} aggregates the result of the best \textit{directed} edge accuracy for three insertion orders of multiple languages (English, French, Italian, Japanese and Chinese).
The results of the top-half (using Zipf-induced frequency) are consistent with the previous experiments: our preferred descent order always achieves the best accuracy across languages with a significant improvement, demonstrating that the core assumption (\textit{Distribution Generality Hypothesis} \cite{weeds2004characterising}) exists in multiple languages and can be leveraged to discover hypernym hierarchy from various word embeddings.

\begin{table}[h]
    \centering
    \begin{tabular}{p{1.5cm}p{1.0cm}p{1.0cm}p{1.1cm}}
    \toprule
    Lang.  &  Descent Order &  Random Order &  Ascend Order  \\
    \hhline{====}
      &\multicolumn{3}{c}{Zipf-induced Frequency}  \\
    \hhline{====}
    $En_{GLoVe}$ &  \textbf{0.0828} &  0.0680 &  0.0482  \\ \hdashline
    $En$ &  \textbf{0.1071} &  0.0780 &  0.0501         \\ \hdashline
    $Fr$ &  \textbf{0.0989} &  0.0459 &  0.0378         \\ \hdashline
    $It$ &  \textbf{0.0843} &  0.0572 &  0.0366         \\ \hdashline
    $Ja$ &  \textbf{0.0752} &  0.0595 &  0.0351         \\ \hdashline
    $Zh$ &  \textbf{0.0907} &  0.0798 &  0.0457         \\ 
    \hhline{====}
    &\multicolumn{3}{c}{PCA-induced Power} \\
    \hhline{====}
    $En_{GLoVe}$ &  \textbf{0.0879} &  0.0705 &  0.0479 \\ \hdashline
    $En$ &  \textbf{0.0839} &  0.0648 &  0.0530 \\ \hdashline
    $Fr$ &  \textbf{0.0970} &  0.0471 &  0.0295 \\ \hdashline
    $It$ &  \textbf{0.078} &  0.0387 &  0.0306 \\ \hdashline
    $Ja$ &  \textbf{0.0711} &  0.0436 &  0.0334 \\ \hdashline
    $Zh$ &  \textbf{0.0959} &  0.0564 &  0.0406 \\

    \bottomrule
    \end{tabular}
    \caption{The best \textit{directed} edge accuracy using cosine distance. The descending order consistently outperforms others. We use GLoVe embedding in $En_{GLoVe}$, and Word2Vec for others.}
    \label{multiwordnet_res}
\end{table}

\begin{figure}[!htbp]
\centering
      \includegraphics[width=5 cm]{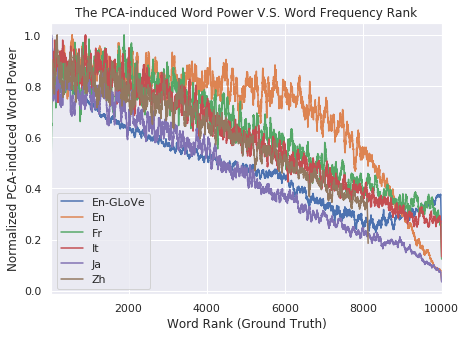}
  \caption{PCA-induced power aligns with word frequency rank, reflecting the word frequency is encoded in the word vector. The value is smoothed using a window of size 50.}
  \label{PCA_freq_vs_freq_rank}
 \end{figure}

Figure \ref{PCA_freq_vs_freq_rank} shows an universal pattern that the PCA-induced word power strongly correlates with the rank of word frequency for all tested word embeddings across different languages. 
Inspired from this interesting observation, we adopt the PCA-induced word power to construct the arborescences over the embeddings that are post-processed as we describes in eq.\ref{pca_equation}.

As we expected, after alternating Zipf-induce frequency to PCA-induced power, the performance is also consistent in Table \ref{multiwordnet_res}, proving that the word vector encodes but only semantic meaning, but also a notion of frequency or generality, leading to the success of hypernym discovery.

\subsection{Least Common Ancestors Discovery}
One of the appealing properties in our proposed tree is least-common-ancestor (LCA) of a word pair $(w_1, w_2)$. We expect LCA can capture the common high level semantic meaning, providing a hierarchical understanding of word pairs. In WordNet, the function \textit{Lowest-Common-Hypernyms} (LCH) plays the similar role. For example, $ LCH(policeman, chef) = {person}$.  

Quantitatively, we evaluate the tree LCA quality by hit-rate, comparing the set $LCA_{tree} (w_1,w_2)$ to the set $LCH_{WordNet} (w_1,w_2)$. If $LCA_{tree}$ contains any element in $LCH_{WordNet}$, we consider it as a successful hit. We also add measure relaxation by including words within two-closure of LCA and LCH.
The hit-rate $s$ is defined as :
\begin{equation}
    \hspace{-1.5cm}
    s = \frac{\sum_{ W^{i} \in W } {\mathbf{1} [ \,  LCA_{tree}(W^{i}) \cap  LCH_{WordNet}(W^{i}) \neq \emptyset ] \,} } 
    {\sum_{ W^{i} \in W } \mathbf{1} [ \, {LCH_{WordNet}(W^{i}) \neq \emptyset } ] \,}  ,
\end{equation}
where $W$ is the set of word pairs. The $i^{th}$ word pair $W^i$ consists of a randomly sampled word $w^i_1$ and another word $w^i_2$ which is sampled from 20-nearest neighbors of $w^i_1$ in the embedding space, making sampled word pair shares semantic meaning with subtle difference. Since the coverage of human annotation in WordNet is limited, we only count for the pairs having non-zero LCH \footnote{In our experiment, 16.64\% of the sampled word pairs have no LCH in English WordNet, 16.29\% for French, 8.58\% for Italian, 0.21\% for Japanese, 23.30\% for Chinese.}. 

We report the best hit-rate for different insertion orders in Table \ref{LCA_hitrate}. As expected, our favored insertion order consistently achieves the highest hit-rate across different languages.

\begin{table}[h]
\centering

\begin{tabular}{p{1.5cm}p{1.1cm}p{1.1cm}p{1.1cm}}
\toprule
 Lang. &  Descent Order &  Random Order &  Ascent Order \\
\hhline{====}
&\multicolumn{3}{c}{Zipf-induced Frequency}  \\
\hhline{====}
  $En_{GLoVe}$ &  \textbf{0.0269} &  0.0136 &  0.0101  \\ \hdashline
  $En$ &  \textbf{0.0385} &  0.0143 &  0.0109\\ \hdashline
  $Fr$ &  \textbf{0.0202} &  0.0010 &  0.0027\\ \hdashline
  $It$ &  \textbf{0.0199} &  0.0031 &  0.0028\\ \hdashline 
  $Ja$ &  \textbf{0.0238} &  0.0038 &  0.0063\\ \hdashline 
  $Zh$ &  \textbf{0.0327} &  0.0081 &  0.0071\\  
\hhline{====}
&\multicolumn{3}{c}{PCA-induced Power}  \\
\hhline{====}
  $En_{GLoVe}$ &  \textbf{0.0299} &  0.0132 &  0.0097\\ \hdashline
  $En$ &  \textbf{0.0151} &  0.0055 &  0.0078\\ \hdashline
  $Fr$ &  \textbf{0.0195} &  0.0075 &  0.0027\\ \hdashline
  $It$ &  \textbf{0.0222} &  0.0115 &  0.0020\\ \hdashline 
  $Ja$ &  \textbf{0.0127} &  0.0020 &  0.0033\\ \hdashline 
  $Zh$ &  \textbf{0.0382} &  0.0055 &  0.0054\\  
\bottomrule
\end{tabular}
\caption{The best tree LCA hit-rate of different insertion order of each language. Our favored descending order consistently outperforms others. We use GLoVe embedding in $En_{GLoVe}$ and Word2Vec (CBOW) for others. 10K word pairs are sampled as mentioned above.  }
\label{LCA_hitrate}
\end{table}

\paragraph{LCA Result Analysis:}
We  dive deep into the LCA results of English (GLoVe) and categorize the results into four cases with examples:
\begin{itemize}
    \item $Success$ case represents the tree LCA overlaps the LCH from WordNet.
        \begin{tcolorbox}[colback=white]
            {\small  
            \makebox[1.5cm][r]{Word Pair :}   \textit{(soup, potato)} \par
            \makebox[1.5cm][r]{LCA :}   \textit{ \{food\} } \par
            \makebox[1.5cm][r]{LCH :}   \textit{ \{food, substance, nutrient ... \} } \par
            }
        \end{tcolorbox}

    \item $Failure_{True}$ case represents the tree LCA fails to overlap LCH. It usually happens when \textit{Distributional Generality} does not hold and should be improved in the future work.
        \begin{tcolorbox}[colback=white]
            {\small  
            \makebox[1.5cm][r]{Word Pair :}  \textit{(september, april)} \par
            \makebox[1.5cm][r]{LCA :}   \textit{ \{june\} } \par
            \makebox[1.5cm][r]{LCH :}   \textit{ \{month, period, measure, ...\} } \par
            }
        \end{tcolorbox}

    \item $Failure_{False}$ case represents the tree LCA fails to overlap LCH, but with rationality.
        
        \begin{tcolorbox}[colback=white]
            {\small  
            \makebox[1.5cm][r]{Word Pair :}  \textit{(thoroughfare, waterway)} \par
            \makebox[1.5cm][r]{LCA :}   \textit{ \{road\} } \par
            \makebox[1.5cm][r]{LCH :}   \textit{ \{physical entity, way\} } \par
            \dotfill \\
            \makebox[1.5cm][r]{Word Pair :}  \textit{(revolt, coup)} \par
            \makebox[1.5cm][r]{LCA :}   \textit{ \{rebellion\} } \par
            \makebox[1.5cm][r]{LCH :}   \textit{ \{group action, event\} } \par

            }
        \end{tcolorbox}

    \item $Complement$ case represents LCH can not be found in WordNet, but our discovered LCA can be complementary. 
        \begin{tcolorbox}[colback=white]
            {\small  
            \makebox[1.5cm][r]{Word Pair :}  \textit{(ford, chevy)} \par
            \makebox[1.5cm][r]{LCA :}   \textit{ \{car\} } \par
            \makebox[1.5cm][r]{LCH :}   \textit{ $\emptyset$ } \par
            \dotfill \\
            \makebox[1.5cm][r]{Word Pair :}  \textit{(convalescence, recuperate)} \par
            \makebox[1.5cm][r]{LCA :}   \textit{ \{recover\} } \par
            \makebox[1.5cm][r]{LCH :}   \textit{ $\emptyset$ } \par
            }

        \end{tcolorbox}

\end{itemize}

\section{Conclusions}
We have proposed an algorithm to construct arborescences from embeddings.
Our experiment shows this arborescence can achieve 62.76\% of directed edge accuracy for Wiki-people graph, and a corresponding 8.98\% and 2.70\% average accuracy in hypernym \& LCA discovery across different languages.
We have restricted attention to directed trees, where each node has only one parent.
Allowing each node to have multiple parents among higher-ranked nodes turns our arborescence into a directed acyclic graph (DAG). 
Such expressibility would allow us to capture alternate word senses, and is left for future work.

\bibliographystyle{acl_natbib}
\bibliography{naacl2021}

\begin{thebibliography}{57}
\expandafter\ifx\csname natexlab\endcsname\relax\def\natexlab#1{#1}\fi

\bibitem[{Alsuhaibani et~al.(2018)Alsuhaibani, Maehara, and
  Bollegala}]{alsuhaibani2018joint}
Mohammed Alsuhaibani, Takanori Maehara, and Danushka Bollegala. 2018.
\newblock Joint learning of hierarchical word embeddings from a corpus and a
  taxonomy.

\bibitem[{Arora et~al.(2016)Arora, Li, Liang, Ma, and
  Risteski}]{arora2016latent}
Sanjeev Arora, Yuanzhi Li, Yingyu Liang, Tengyu Ma, and Andrej Risteski. 2016.
\newblock A latent variable model approach to pmi-based word embeddings.
\newblock \emph{Transactions of the Association for Computational Linguistics},
  4:385--399.

\bibitem[{Bender et~al.(2001)Bender, Pemmasani, Skiena, and
  Sumazin}]{bender2001finding}
Michael~A Bender, Giridhar Pemmasani, Steven Skiena, and Pavel Sumazin. 2001.
\newblock Finding least common ancestors in directed acyclic graphs.
\newblock In \emph{Proceedings of the twelfth annual ACM-SIAM symposium on
  Discrete algorithms}, pages 845--854. Society for Industrial and Applied
  Mathematics.

\bibitem[{Bond and Foster(2013)}]{bond2013linking}
Francis Bond and Ryan Foster. 2013.
\newblock Linking and extending an open multilingual wordnet.
\newblock In \emph{Proceedings of the 51st Annual Meeting of the Association
  for Computational Linguistics (Volume 1: Long Papers)}, pages 1352--1362.

\bibitem[{Bond and Paik(2012)}]{bond2012survey}
Francis Bond and Kyonghee Paik. 2012.
\newblock A survey of wordnets and their licenses.
\newblock \emph{Small}, 8(4):5.

\bibitem[{Borca-Tasciuc et~al.(2022)Borca-Tasciuc, Guo, Bak, and
  Skiena}]{borca2022provable}
Giorgian Borca-Tasciuc, Xingzhi Guo, Stanley Bak, and Steven Skiena. 2022.
\newblock Provable fairness for neural network models using formal
  verification.
\newblock \emph{arXiv preprint arXiv:2212.08578}.

\bibitem[{Chen et~al.(2023)Chen, Guo, Zhou, Yang, and
  Skiena}]{chen2023accelerating}
Zhen Chen, Xingzhi Guo, Baojian Zhou, Deqing Yang, and Steven Skiena. 2023.
\newblock Accelerating personalized pagerank vector computation.
\newblock In \emph{Proceedings of the 29th ACM SIGKDD Conference on Knowledge
  Discovery and Data Mining}, pages 262--273.

\bibitem[{Fellbaum(1998)}]{_Fellbaum:1998}
Christiane Fellbaum, editor. 1998.
\newblock MIT Press, Cambridge, MA.

\bibitem[{Fu et~al.(2014)Fu, Guo, Qin, Che, Wang, and Liu}]{fu2014learning}
Ruiji Fu, Jiang Guo, Bing Qin, Wanxiang Che, Haifeng Wang, and Ting Liu. 2014.
\newblock Learning semantic hierarchies via word embeddings.
\newblock In \emph{Proceedings of the 52nd Annual Meeting of the Association
  for Computational Linguistics (Volume 1: Long Papers)}, pages 1199--1209.

\bibitem[{Gaume et~al.(2006)Gaume, Venant, and Victorri}]{gaume2006hierarchy}
Bruno Gaume, Fabienne Venant, and Bernard Victorri. 2006.
\newblock Hierarchy in lexical organisation of natural languages.
\newblock In \emph{Hierarchy in natural and social sciences}, pages 121--142.
  Springer.

\bibitem[{Gillespie et~al.(2020)Gillespie, Konstantakopoulos, Guo, Vasudevan,
  and Sethy}]{gillespie2020improving}
Kellen Gillespie, Ioannis~C Konstantakopoulos, Xingzhi Guo, Vishal~Thanvantri
  Vasudevan, and Abhinav Sethy. 2020.
\newblock Improving device directedness classification of utterances with
  semantic lexical features.
\newblock In \emph{ICASSP 2020-2020 IEEE International Conference on Acoustics,
  Speech and Signal Processing (ICASSP)}, pages 7859--7863. IEEE.

\bibitem[{Gladkova et~al.(2016)Gladkova, Drozd, and
  Matsuoka}]{gladkovaetal2016analogy}
Anna Gladkova, Aleksandr Drozd, and Satoshi Matsuoka. 2016.
\newblock \href {https://doi.org/10.18653/v1/N16-2002} {Analogy-based detection
  of morphological and semantic relations with word embeddings: what works and
  what doesn{'}t.}
\newblock In \emph{Proceedings of the {NAACL} Student Research Workshop}, pages
  8--15, San Diego, California. Association for Computational Linguistics.

\bibitem[{Gong et~al.(2018)Gong, He, Tan, Qin, Wang, and Liu}]{gong2018frage}
Chengyue Gong, Di~He, Xu~Tan, Tao Qin, Liwei Wang, and Tie-Yan Liu. 2018.
\newblock Frage: Frequency-agnostic word representation.
\newblock In \emph{Advances in neural information processing systems}, pages
  1334--1345.

\bibitem[{Guo et~al.(2019)Guo, Huang, Gamborino, Tseng, Fu, and
  Yeh}]{guo2019inferring}
Xingzhi Guo, Yu-Cian Huang, Edwinn Gamborino, Shih-Huan Tseng, Li-Chen Fu, and
  Su-Ling Yeh. 2019.
\newblock Inferring human feelings and desires for human-robot trust promotion.
\newblock In \emph{International Conference on Human-Computer Interaction},
  pages 365--375. Springer.

\bibitem[{Guo et~al.(2022{\natexlab{a}})Guo, Kondracki, Nikiforakis, and
  Skiena}]{guo2022verba}
Xingzhi Guo, Brian Kondracki, Nick Nikiforakis, and Steven Skiena.
  2022{\natexlab{a}}.
\newblock Verba volant, scripta volant: Understanding post-publication title
  changes in news outlets.
\newblock In \emph{Proceedings of the ACM Web Conference 2022}, pages 588--598.

\bibitem[{Guo et~al.(2021)Guo, Zhou, and Skiena}]{guo2021subset}
Xingzhi Guo, Baojian Zhou, and Steven Skiena. 2021.
\newblock Subset node representation learning over large dynamic graphs.
\newblock In \emph{Proceedings of the 27th ACM SIGKDD Conference on Knowledge
  Discovery \& Data Mining}, pages 516--526.

\bibitem[{Guo et~al.(2022{\natexlab{b}})Guo, Zhou, and Skiena}]{guo2022subset}
Xingzhi Guo, Baojian Zhou, and Steven Skiena. 2022{\natexlab{b}}.
\newblock Subset node anomaly tracking over large dynamic graphs.
\newblock In \emph{Proceedings of the 28th ACM SIGKDD Conference on Knowledge
  Discovery and Data Mining}, pages 475--485.

\bibitem[{Harris(1954)}]{harris1954distributional}
Zellig~S Harris. 1954.
\newblock Distributional structure.
\newblock \emph{Word}, 10(2-3):146--162.

\bibitem[{Hearst(1992)}]{hearst1992automatic}
Marti~A Hearst. 1992.
\newblock Automatic acquisition of hyponyms from large text corpora.
\newblock In \emph{Proceedings of the 14th conference on Computational
  linguistics-Volume 2}, pages 539--545. Association for Computational
  Linguistics.

\bibitem[{Herbelot and Ganesalingam(2013)}]{herbelot2013measuring}
Aur{\'e}lie Herbelot and Mohan Ganesalingam. 2013.
\newblock Measuring semantic content in distributional vectors.
\newblock In \emph{Proceedings of the 51st Annual Meeting of the Association
  for Computational Linguistics (Volume 2: Short Papers)}, pages 440--445.

\bibitem[{Isahara et~al.(2008)Isahara, Bond, Uchimoto, Utiyama, and
  Kanzaki}]{Isahara:Bond:Uchimoto:Utiyama:Kanzaki:2008}
Hitoshi Isahara, Francis Bond, Kiyotaka Uchimoto, Masao Utiyama, and Kyoko
  Kanzaki. 2008.
\newblock Development of the {Japanese} {WordNet}.
\newblock In \emph{Sixth International conference on Language Resources and
  Evaluation (LREC 2008)}, Marrakech.

\bibitem[{Leskovec and Krevl(2014)}]{snapnets}
Jure Leskovec and Andrej Krevl. 2014.
\newblock {SNAP Datasets}: {Stanford} large network dataset collection.
\newblock \url{http://snap.stanford.edu/data}.

\bibitem[{Li(1992)}]{li1992random}
Wentian Li. 1992.
\newblock Random texts exhibit zipf's-law-like word frequency distribution.
\newblock \emph{IEEE Transactions on information theory}, 38(6):1842--1845.

\bibitem[{Lin et~al.(2023)Lin, Feng, Guo, Zhang, Yin, Kwoh, and
  Xu}]{lin2023comet}
Zhuoyi Lin, Lei Feng, Xingzhi Guo, Yu~Zhang, Rui Yin, Chee~Keong Kwoh, and Chi
  Xu. 2023.
\newblock Comet: Convolutional dimension interaction for collaborative
  filtering.
\newblock \emph{ACM Transactions on Intelligent Systems and Technology},
  14(4):1--18.

\bibitem[{Mikolov et~al.(2013{\natexlab{a}})Mikolov, Chen, Corrado, and
  Dean}]{mikolov2013efficient}
Tomas Mikolov, Kai Chen, Greg Corrado, and Jeffrey Dean. 2013{\natexlab{a}}.
\newblock Efficient estimation of word representations in vector space.
\newblock \emph{arXiv preprint arXiv:1301.3781}.

\bibitem[{Mikolov et~al.(2013{\natexlab{b}})Mikolov, Sutskever, Chen, Corrado,
  and Dean}]{mikolov2013distributed}
Tomas Mikolov, Ilya Sutskever, Kai Chen, Greg~S Corrado, and Jeff Dean.
  2013{\natexlab{b}}.
\newblock Distributed representations of words and phrases and their
  compositionality.
\newblock In \emph{Advances in neural information processing systems}, pages
  3111--3119.

\bibitem[{Miller(1995)}]{miller1995wordnet}
George~A Miller. 1995.
\newblock Wordnet: a lexical database for english.
\newblock \emph{Communications of the ACM}, 38(11):39--41.

\bibitem[{Mu et~al.(2017)Mu, Bhat, and Viswanath}]{mu2017all}
Jiaqi Mu, Suma Bhat, and Pramod Viswanath. 2017.
\newblock All-but-the-top: Simple and effective postprocessing for word
  representations.
\newblock \emph{arXiv preprint arXiv:1702.01417}.

\bibitem[{Navigli and Ponzetto(2012)}]{navigli2012babelnet}
Roberto Navigli and Simone~Paolo Ponzetto. 2012.
\newblock Babelnet: The automatic construction, evaluation and application of a
  wide-coverage multilingual semantic network.
\newblock \emph{Artificial Intelligence}, 193:217--250.

\bibitem[{Nickel and Kiela(2017)}]{nickel2017poincare}
Maximillian Nickel and Douwe Kiela. 2017.
\newblock Poincar{\'e} embeddings for learning hierarchical representations.
\newblock In \emph{Advances in neural information processing systems}, pages
  6338--6347.

\bibitem[{Omohundro(1989)}]{omohundro1989five}
Stephen~M Omohundro. 1989.
\newblock \emph{Five balltree construction algorithms}.
\newblock International Computer Science Institute Berkeley.

\bibitem[{Page et~al.(1999)Page, Brin, Motwani, and
  Winograd}]{page1999pagerank}
Lawrence Page, Sergey Brin, Rajeev Motwani, and Terry Winograd. 1999.
\newblock The pagerank citation ranking: Bringing order to the web.
\newblock Technical report, Stanford InfoLab.

\bibitem[{Pennington et~al.(2014)Pennington, Socher, and
  Manning}]{pennington2014glove}
Jeffrey Pennington, Richard Socher, and Christopher Manning. 2014.
\newblock Glove: Global vectors for word representation.
\newblock In \emph{Proceedings of the 2014 conference on empirical methods in
  natural language processing (EMNLP)}, pages 1532--1543.

\bibitem[{Perozzi et~al.(2014)Perozzi, Al-Rfou, and
  Skiena}]{perozzi2014deepwalk}
Bryan Perozzi, Rami Al-Rfou, and Steven Skiena. 2014.
\newblock Deepwalk: Online learning of social representations.
\newblock In \emph{Proceedings of the 20th ACM SIGKDD international conference
  on Knowledge discovery and data mining}, pages 701--710. ACM.

\bibitem[{Polguere(2009)}]{polguere2009lexical}
Alain Polguere. 2009.
\newblock Lexical systems: graph models of natural language lexicons.
\newblock \emph{Language resources and evaluation}, 43(1):41--55.

\bibitem[{Rimell(2014)}]{rimell2014distributional}
Laura Rimell. 2014.
\newblock Distributional lexical entailment by topic coherence.
\newblock In \emph{Proceedings of the 14th Conference of the European Chapter
  of the Association for Computational Linguistics}, pages 511--519.

\bibitem[{Sagot and Fi\v{s}er(2008)}]{Sagot:Fiser:2008}
Beno\^{i}t Sagot and Darja Fi\v{s}er. 2008.
\newblock Building a free {French} wordnet from multilingual resources.
\newblock In \emph{Proceedings of the Sixth International Language Resources
  and Evaluation (LREC'08)}, Marrakech, Morocco.

\bibitem[{Schieber and Vishkin(1988)}]{schieber1988finding}
Baruch Schieber and Uzi Vishkin. 1988.
\newblock On finding lowest common ancestors: Simplification and
  parallelization.
\newblock \emph{SIAM Journal on Computing}, 17(6):1253--1262.

\bibitem[{Schnabel et~al.(2015)Schnabel, Labutov, Mimno, and
  Joachims}]{schnabel2015evaluation}
Tobias Schnabel, Igor Labutov, David Mimno, and Thorsten Joachims. 2015.
\newblock Evaluation methods for unsupervised word embeddings.
\newblock In \emph{Proceedings of the 2015 Conference on Empirical Methods in
  Natural Language Processing}, pages 298--307.

\bibitem[{Skiena and Ward(2014)}]{skiena2014s}
Steven Skiena and Charles~B Ward. 2014.
\newblock \emph{Who's bigger?: Where historical figures really rank}.
\newblock Cambridge University Press.

\bibitem[{Snow et~al.(2005)Snow, Jurafsky, and Ng}]{snow2005learning}
Rion Snow, Daniel Jurafsky, and Andrew~Y Ng. 2005.
\newblock Learning syntactic patterns for automatic hypernym discovery.
\newblock In \emph{Advances in neural information processing systems}, pages
  1297--1304.

\bibitem[{Sultan et~al.(2022)Sultan, Guo, and Skiena}]{sultan2022low}
Syed~Fahad Sultan, Xingzhi Guo, and Steven Skiena. 2022.
\newblock Low-dimensional genotype embeddings for predictive models.
\newblock In \emph{Proceedings of the 13th ACM International Conference on
  Bioinformatics, Computational Biology and Health Informatics}, pages 1--4.

\bibitem[{Tifrea et~al.(2018)Tifrea, B{\'e}cigneul, and
  Ganea}]{tifrea2018poincar}
Alexandru Tifrea, Gary B{\'e}cigneul, and Octavian-Eugen Ganea. 2018.
\newblock Poincar$\backslash$'e glove: Hyperbolic word embeddings.
\newblock \emph{arXiv preprint arXiv:1810.06546}.

\bibitem[{Toral et~al.(2010)Toral, Bracale, Monachini, and
  Soria}]{Toral:Bracale:Monachini:Soria:2010}
Antonio Toral, Stefania Bracale, Monica Monachini, and Claudia Soria. 2010.
\newblock Rejuvenating the italian wordnet: upgrading, standardising,
  extending.
\newblock In \emph{Proceedings of the 5th International Conference of the
  Global WordNet Association (GWC-2010)}, Mumbai.

\bibitem[{Vylomova et~al.(2015)Vylomova, Rimell, Cohn, and
  Baldwin}]{vylomova2015take}
Ekaterina Vylomova, Laura Rimell, Trevor Cohn, and Timothy Baldwin. 2015.
\newblock Take and took, gaggle and goose, book and read: Evaluating the
  utility of vector differences for lexical relation learning.
\newblock \emph{arXiv preprint arXiv:1509.01692}.

\bibitem[{Wang and Bond(2013)}]{Wang:Bond:2013}
Shan Wang and Francis Bond. 2013.
\newblock Building the chinese open wordnet (cow): Starting from core synsets.
\newblock In \emph{Sixth International Joint Conference on Natural Language
  Processing}, pages 10--18.

\bibitem[{Wang et~al.(2019)Wang, Jin, Wang, Cui, Ma, and
  Qu}]{wang2019deepdrawing}
Yong Wang, Zhihua Jin, Qianwen Wang, Weiwei Cui, Tengfei Ma, and Huamin Qu.
  2019.
\newblock Deepdrawing: A deep learning approach to graph drawing.
\newblock \emph{IEEE transactions on visualization and computer graphics}.

\bibitem[{Wang(2023)}]{wang2023knowledge2}
Yuheng Wang. 2023.
\newblock \emph{Knowledge Authoring With Factual English, Rules, and Actions}.
\newblock Ph.D. thesis, State University of New York at Stony Brook.

\bibitem[{Wang et~al.(2022)Wang, Borca-Tasciuc, Goel, Fodor, and
  Kifer}]{wang2022knowledge}
Yuheng Wang, Giorgian Borca-Tasciuc, Nikhil Goel, Paul Fodor, and Michael
  Kifer. 2022.
\newblock Knowledge authoring with factual english.
\newblock \emph{arXiv preprint arXiv:2208.03094}.

\bibitem[{Wang et~al.(2023)Wang, Fodor, and Kifer}]{wang2023knowledge}
Yuheng Wang, Paul Fodor, and Michael Kifer. 2023.
\newblock Knowledge authoring for rules and actions.
\newblock \emph{Theory and Practice of Logic Programming}, 23(4):797--811.

\bibitem[{Weeds et~al.(2014)Weeds, Clarke, Reffin, Weir, and
  Keller}]{weeds2014learning}
Julie Weeds, Daoud Clarke, Jeremy Reffin, David Weir, and Bill Keller. 2014.
\newblock Learning to distinguish hypernyms and co-hyponyms.
\newblock In \emph{Proceedings of COLING 2014, the 25th International
  Conference on Computational Linguistics: Technical Papers}, pages 2249--2259.
  Dublin City University and Association for Computational Linguistics.

\bibitem[{Weeds et~al.(2004)Weeds, Weir, and
  McCarthy}]{weeds2004characterising}
Julie Weeds, David Weir, and Diana McCarthy. 2004.
\newblock Characterising measures of lexical distributional similarity.
\newblock In \emph{Proceedings of the 20th international conference on
  Computational Linguistics}, page 1015. Association for Computational
  Linguistics.

\bibitem[{Zhang et~al.(2023)Zhang, Xiang, Guo, Zhou, and
  Yang}]{zhang2023subanom}
Chi Zhang, Wenkai Xiang, Xingzhi Guo, Baojian Zhou, and Deqing Yang. 2023.
\newblock Subanom: Efficient subgraph anomaly detection framework over dynamic
  graphs.
\newblock In \emph{2023 IEEE International Conference on Data Mining Workshops
  (ICDMW)}, pages 1178--1185. IEEE.

\bibitem[{Zhang et~al.(2016)Zhang, Wang, Wang, and Zhou}]{zhang2016symmetrical}
Junqi Zhang, Yuheng Wang, Cheng Wang, and Mengchu Zhou. 2016.
\newblock Symmetrical hierarchical stochastic searching on the line in
  informative and deceptive environments.
\newblock \emph{IEEE transactions on cybernetics}, 47(3):626--635.

\bibitem[{Zhang et~al.(2017)Zhang, Wang, and Zhou}]{zhang2017fast2}
JunQi Zhang, YuHeng Wang, and MengChu Zhou. 2017.
\newblock Fast adaptive search on the line in dual environments.
\newblock In \emph{2017 13th IEEE Conference on Automation Science and
  Engineering (CASE)}, pages 1540--1545. IEEE.

\bibitem[{Zhang et~al.(2018)Zhang, Zhu, Wang, and Zhou}]{zhang2018dual}
Junqi Zhang, Xixun Zhu, Yuheng Wang, and MengChu Zhou. 2018.
\newblock Dual-environmental particle swarm optimizer in noisy and noise-free
  environments.
\newblock \emph{IEEE transactions on cybernetics}, 49(6):2011--2021.

\bibitem[{Zhou et~al.(2024)Zhou, Sun, Harikandeh, Guo, Yang, and
  Xiao}]{zhou2024iterative}
Baojian Zhou, Yifan Sun, Reza~Babanezhad Harikandeh, Xingzhi Guo, Deqing Yang,
  and Yanghua Xiao. 2024.
\newblock Iterative methods via locally evolving set process.
\newblock \emph{arXiv preprint arXiv:2410.15020}.

\end{thebibliography}

\end{document}